\documentclass[conference]{IEEEtran}
\IEEEoverridecommandlockouts

\usepackage{amsmath,amssymb,amsfonts}
\usepackage{graphicx}
\usepackage{textcomp}
\usepackage{xcolor}
\usepackage{booktabs}
\usepackage{array}
\usepackage{url}
\usepackage[hidelinks]{hyperref}
\usepackage{cite}
\usepackage{tikz}
\usepackage{balance}
\usetikzlibrary{positioning,calc}
\usepackage{pgfplots}
\pgfplotsset{compat=1.18}

\def\BibTeX{{\rm B\kern-.05em{\sc i\kern-.025em b}\kern-.08em T\kern-.1667em\lower.7ex\hbox{E}\kern-.125emX}}

\begin{document}

\title{Combining Trained Models in Reinforcement Learning\thanks{This work was conducted as part of the MSc Autonomous Systems programme at Hochschule Bonn-Rhein-Sieg (H-BRS), Germany.}}

\author{
\IEEEauthorblockN{Ujjwal Patil}
\IEEEauthorblockA{\textit{Department of Computer Science}\\
\textit{Hochschule Bonn-Rhein-Sieg (H-BRS)}\\
Sankt Augustin, Germany\\
ujjwal.patil@smail.inf.h-brs.de}
\and
\IEEEauthorblockN{Javad Ghofrani}
\IEEEauthorblockA{\textit{Department of Computer Science}\\
\textit{Hochschule Bonn-Rhein-Sieg (H-BRS)}\\
Sankt Augustin, Germany\\
javad.ghofrani@h-brs.de}
}

\maketitle

\begin{abstract}
Deep reinforcement learning (DRL) has delivered strong results in domains such as Atari and Go, but it still suffers from high sample cost and weak transfer beyond the training setting \cite{Mnih2015,Silver2017}. A common response is to reuse information from previously trained models through transfer, distillation, ensemble methods, or federated training instead of learning each target task from random initialization. The literature on these mechanisms is fragmented, and published comparisons are hard to interpret because tasks, baselines, and compute budgets differ.

This paper presents a PRISMA-guided systematic review of empirical studies on pretrained knowledge reuse in DRL. Starting from 589 records retrieved from IEEE Xplore, the ACM Digital Library, and citation tracing, we screened 570 unique records and assessed 89 full texts. After applying the final eligibility criteria, 15 empirical studies remained in the main synthesis. We analyzed them qualitatively across three factors: source-target similarity, diversity among reused models, and the fairness of comparisons against from-scratch baselines.

Three patterns recur across the surviving corpus. First, positive results are concentrated in settings where source and target tasks share substantial structure or where the method includes an explicit gating or alignment mechanism. Second, evidence for ensembles and federated aggregation is promising but sparse and mostly limited to narrow settings. Third, compute-matched comparisons are rare, which weakens claims about efficiency gains over stronger single-agent baselines. The paper contributes a narrower and internally consistent review scope, a study-level synthesis of empirical evidence, and a provisional independence spectrum that should be treated as a hypothesis for future benchmarking rather than a validated metric.
\end{abstract}


\section{Introduction}

Deep reinforcement learning (DRL) has shown that neural policies can master challenging sequential decision problems, from Atari games to Go \cite{Mnih2015,Silver2017}. Those headline results hide two practical weaknesses. Training is often sample-hungry, and policies that perform well on one task or environment usually degrade when the task, observation process, or dynamics shift.

One response is to reuse information from previously trained models rather than restart training from scratch for every target problem. In DRL, this reuse appears in several forms: distilling a teacher into a student \cite{Rusu2016}, reusing policies across related tasks \cite{Li2019,Liu2024}, combining multiple models to stabilize estimation \cite{Park2024}, and aggregating updates across decentralized agents \cite{Sethi2023,An2025}. These lines of work are usually discussed in separate subfields. That fragmentation makes it difficult to answer a simple question: under what conditions does pretrained knowledge reuse improve over training a single agent from scratch on the target task?

This review addresses that question through a qualitative synthesis of empirical DRL studies. The paper does not claim a unified experimental benchmark or a formal meta-analysis. The underlying studies use different tasks, metrics, model classes, and baselines, so pooled effect sizes would not be credible. Instead, the goal is to map what has been tested, identify recurring conditions behind positive and negative findings, and make the limitations of the current evidence explicit.

The review makes three concrete contributions. First, it defines the scope around pretrained knowledge reuse in DRL. Second, it synthesizes the empirical evidence across transfer, distillation, ensemble, and federated settings using a common conceptual comparator: performance relative to a target-task baseline trained without reused pretrained knowledge. Third, it proposes an independence spectrum as a provisional lens for describing how reused models differ in training experience. The spectrum is presented as a hypothesis generated from the literature, not as a validated explanatory law.

\section{Related Work and Review Scope}
\label{sec:rw}

Prior reviews have examined transfer reinforcement learning, sim-to-real adaptation, or single sub-areas in isolation. Taylor and Stone \cite{Taylor2009} remain the standard survey on transfer in reinforcement learning, but the paper predates modern deep RL. Zhao et al.\ \cite{Zhao2020} review sim-to-real DRL for robotics, and Wei et al.\ \cite{Wei2025} survey transfer RL more broadly. Those papers are useful background, but they are not cross-mechanism empirical syntheses of pretrained model reuse in DRL.

This review draws a sharper boundary around the intervention. A study is in scope only if reused pretrained model information is a central part of the method and the paper reports empirical DRL results. Pure surveys, hardware-throughput papers, and inference-only filtering or defense methods are treated as adjacent work, not as evidence for the main question. This matters because a review becomes incoherent fast when it mixes actual knowledge reuse during learning with papers that only speed up simulation or combine outputs after training.

Within that narrower scope, four mechanism families recur:
\begin{itemize}
\item \textbf{Distillation}, where a teacher policy or value function supervises a student model \cite{Rusu2016,Qu2022,Yu2024}.
\item \textbf{Transfer and policy reuse}, where previously learned policies, skills, or priors are reused on a target task \cite{Li2019,Garcia2018,Zhuang2022,Liu2024,Zhang2023,Du2025}.
\item \textbf{Ensembles}, where multiple learned models jointly influence control or value estimation \cite{Park2024}.
\item \textbf{Federated training}, where multiple agents train locally and aggregate updates without centralizing raw data \cite{Sethi2023,An2025}.
\end{itemize}

Recent adjacent work also explores feature-level combination of multiple pretrained encoders for RL \cite{Piccoli2025}. Because that paper appeared as a 2025 workshop/preprint rather than an archival venue, it is discussed as background but excluded from the main synthesis.

\section{Methods}
\label{sec:methods}

\subsection{Review design}

The review was conducted as a PRISMA-guided systematic review and uses the PRISMA 2020 reporting standard as its methodological reference \cite{Page2021PRISMA}. Screening, extraction, and coding were performed by a single reviewer. That choice makes the workflow feasible for a master's thesis project, but it also increases the risk of selection and interpretation bias. The resulting synthesis should therefore be read as a careful single-reviewer evidence map, not as a fully duplicated systematic review.

\subsection{Search sources and strategy}

The search covered IEEE Xplore, the ACM Digital Library, and forward/backward citation tracing from eligible papers. The original search was conducted during August 2025. The day-level search log was not preserved in the initial project materials, which limits exact reproducibility. To compensate, the final manuscript reports the search logic explicitly.

\begin{sloppypar}
The search combined an intervention block with an RL-context block. The intervention block used the terms \texttt{"policy distillation"}, \texttt{"knowledge distillation"}, \texttt{"policy reuse"}, \texttt{"policy transfer"}, \texttt{"transfer reinforcement learning"}, \texttt{"ensemble reinforcement learning"}, \texttt{"federated reinforcement learning"}, \texttt{"model merging"}, and \texttt{"weight averaging"}. The RL-context block used \texttt{"reinforcement learning"}, \texttt{"deep reinforcement learning"}, \texttt{"actor-critic"}, \texttt{"policy gradient"}, \texttt{"deep Q-network"}, and \texttt{"DQN"}. Searches were restricted to title and abstract fields to avoid large numbers of irrelevant hardware-parallelism papers.
\end{sloppypar}

Table \ref{tab:search} reports the retrieval and screening counts. Exact query templates are provided in Appendix~\ref{app:queries}.

\begin{table}[htbp]
\caption{Search and screening summary}
\label{tab:search}
\renewcommand{\arraystretch}{1.1}
\begin{center}
\small
\begin{tabular}{lrrr}
\toprule
\textbf{Source} & \textbf{Initial} & \textbf{Post-dedup} & \textbf{Main synthesis} \\
\midrule
IEEE Xplore & 547 & 541 & 11 \\
ACM Digital Library & 30 & 27 & 3 \\
Citation tracing & 12 & 12 & 1 \\
\midrule
\textbf{Total} & \textbf{589} & \textbf{570} & \textbf{15} \\
\bottomrule
\end{tabular}
\end{center}
\end{table}

\subsection{Eligibility criteria}

The final eligibility criteria were tightened relative to the earlier framing because the first version mixed empirical studies with adjacent but ineligible material. Studies were included in the main synthesis if they satisfied all four conditions below:
\begin{enumerate}
\item They reported an empirical DRL study published between 2015 and 2025.
\item Reuse of pretrained model information was a central method component.
\item They compared the proposed method against at least one meaningful baseline on task performance, sample efficiency, robustness, or transfer.
\item The paper appeared in an archival peer-reviewed venue, or in the case of foundational early work, in a peer-reviewed conference track widely treated as part of the field's core literature.
\end{enumerate}

The main synthesis excluded surveys, tutorials, opinion pieces, pure hardware-throughput papers, and methods that only filter or combine outputs at inference without affecting representation learning, policy learning, or parameter updates. Table \ref{tab:excluded_adjacent} in Appendix~\ref{app:adjacent} lists adjacent studies that were identified during screening but excluded from the main empirical synthesis.

\subsection{Selection process}

\begin{figure*}[t]
\centering
\begin{tikzpicture}[
node distance = 1.05cm and 2.2cm,
main/.style = {
rectangle, draw=black, line width=0.7pt,
text width=7.7cm, align=center,
font=\small, inner sep=7pt, minimum height=1.1cm
},
side/.style = {
rectangle, draw=black, line width=0.7pt,
text width=4.1cm, align=center,
font=\small, inner sep=6pt, minimum height=1.0cm
},
phase/.style = {
rectangle, draw=black, fill=gray!18, line width=0.7pt,
text width=1.7cm, align=center,
font=\small\bfseries, inner sep=5pt, minimum height=1.1cm
},
arr/.style = {->, thick, >=stealth},
darr/.style = {->, thick, >=stealth, dashed},
]
\node[main] (ID) {%
\textbf{Records identified}\\[2pt]
IEEE Xplore:\ n\,=\,547 \quad
ACM Digital Library:\ n\,=\,30\\
Citation tracing:\ n\,=\,12\\[3pt]
\textbf{Total:\ n\,=\,589}};
\node[main, below=of ID] (SCR) {%
\textbf{Records screened}\\[2pt]
(19 duplicates removed)\\[2pt]
\textbf{n\,=\,570}};
\node[main, below=of SCR] (ELI) {%
\textbf{Full-text articles assessed}\\[2pt]
\textbf{n\,=\,89}};
\node[main, below=of ELI, fill=gray!8] (INC) {%
\textbf{Empirical studies included in main synthesis}\\[2pt]
\textbf{n\,=\,15}};
\node[side, right=of SCR] (EXCL1) {%
\textbf{Records excluded}\\
at title/abstract stage\\[2pt]
n\,=\,481};
\node[side, right=of ELI] (EXCL2) {%
\textbf{Full texts excluded}\\
(surveys, non-empirical, inference-only,\\
hardware-throughput, out of scope)\\[2pt]
n\,=\,74};
\node[phase, left=1.6cm of ID] (P1) {Identification};
\node[phase, left=1.6cm of SCR] (P2) {Screening};
\node[phase, left=1.6cm of ELI] (P3) {Eligibility};
\node[phase, left=1.6cm of INC] (P4) {Included};
\draw[arr] (ID.south) -- (SCR.north);
\draw[arr] (SCR.south) -- (ELI.north);
\draw[arr] (ELI.south) -- (INC.north);
\draw[darr] (SCR.east) -- (EXCL1.west);
\draw[darr] (ELI.east) -- (EXCL2.west);
\draw[gray!50, thin] (P1.south) -- (P2.north);
\draw[gray!50, thin] (P2.south) -- (P3.north);
\draw[gray!50, thin] (P3.south) -- (P4.north);
\end{tikzpicture}
\caption{PRISMA-style study flow for the final main synthesis.}
\label{fig:prisma}
\end{figure*}
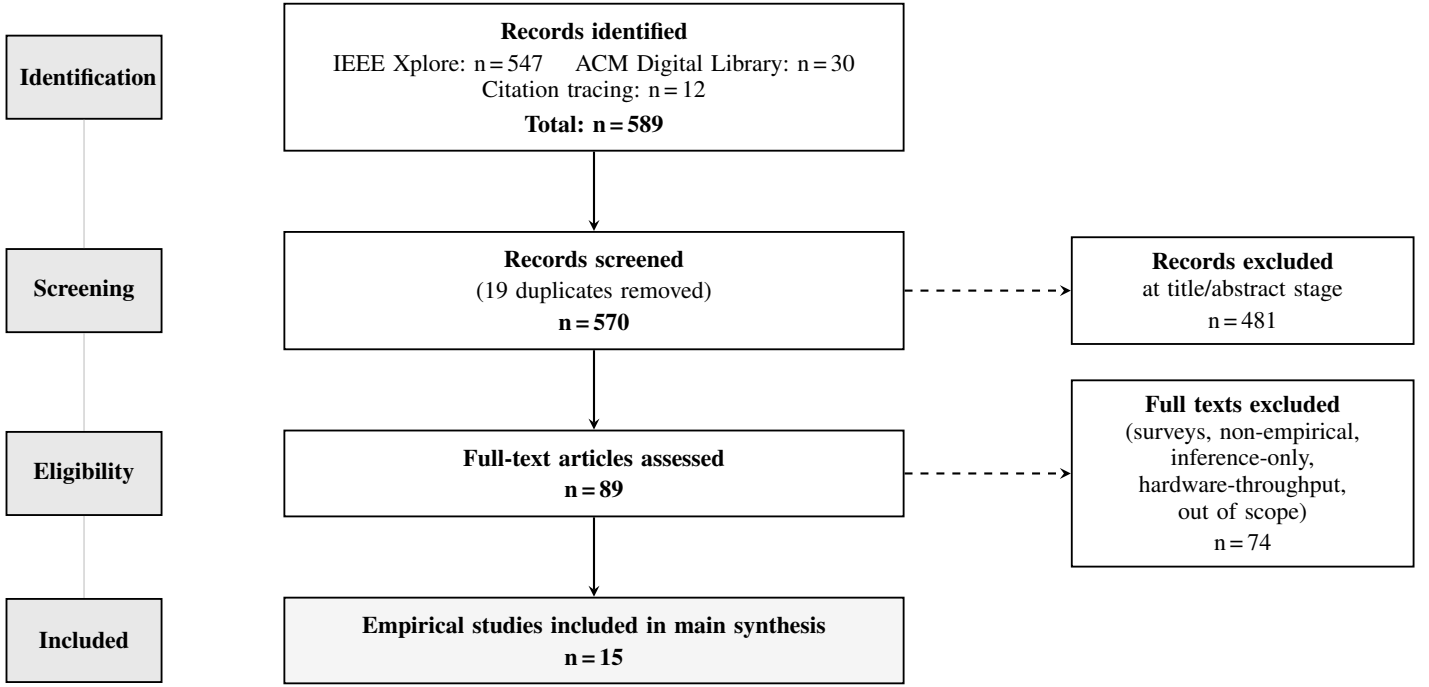

The search retrieved 589 records. After removing 19 duplicates, 570 unique records remained. Title and abstract screening reduced the pool to 89 full texts. A final eligibility audit excluded 74 full-text papers, leaving 15 empirical studies in the main synthesis. Figure \ref{fig:prisma} shows the final flow.

\subsection{Data extraction and evidence audit}

The extraction scheme records the following items for each included study: environment or task family, reuse mechanism, number of reused models, source-target relation, baseline type, reported outcome, and any statement about compute or uncertainty.

Instead of MMAT, this review uses a lightweight evidence audit better suited to ML systems papers. Each study was assessed on five reporting dimensions:
\begin{enumerate}
\item clarity of the task and environment,
\item clarity of the comparison baseline,
\item whether uncertainty or variance across runs was reported,
\item whether compute or training-budget information was reported,
\item whether the relation between source and target tasks was made explicit.
\end{enumerate}
The audit was used to qualify confidence in the synthesis, not as an exclusion threshold.

\subsection{Synthesis strategy}

Because the included studies use different environments, objectives, and metrics, the review uses qualitative narrative synthesis rather than meta-analysis. Three cross-cutting factors were coded during synthesis:
\begin{enumerate}
\item \textbf{source-target similarity}, meaning shared dynamics, representation, reward structure, or task family;
\item \textbf{model independence}, meaning how reused models differ in seed, data, task, or reward;
\item \textbf{comparison fairness}, especially whether the from-scratch baseline was plausibly matched on training budget.
\end{enumerate}

\section{Results}
\label{sec:results}

\subsection{Overview of the included corpus}

The final main synthesis contains 15 empirical studies. Transfer and distillation dominate the evidence base, accounting for 12 of the 15 included papers. Only one paper tests an ensemble mechanism that directly influences learning dynamics in the final control loop \cite{Park2024}, and only two papers study federated reinforcement learning \cite{Sethi2023,An2025}. That imbalance matters: strong claims about cross-mechanism differences would be fake precision, because the evidence is concentrated in transfer-style settings.

\begin{figure}[htbp]
\centering
\begin{tikzpicture}
\begin{axis}[
width=\columnwidth,
height=3.7cm,
ybar stacked,
bar width=10pt,
xlabel={Year},
ylabel={\# Papers},
xtick={2015,2018,2019,2022,2023,2024,2025},
xticklabel style={rotate=45, anchor=east, font=\tiny},
yticklabel style={font=\tiny},
xlabel style={font=\small},
ylabel style={font=\small},
legend style={font=\tiny, at={(0.5,-0.45)}, anchor=north, legend columns=3},
ymin=0, ymax=4,
ymajorgrids=true,
grid style=dashed,
enlarge x limits=0.08,
]
\addplot+[ybar, fill=blue!60, draw=blue!80] coordinates {
(2015,1)(2018,2)(2019,2)(2022,2)(2023,1)(2024,2)(2025,2)};
\addplot+[ybar, fill=orange!70, draw=orange!90] coordinates {
(2015,0)(2018,0)(2019,0)(2022,0)(2023,0)(2024,1)(2025,0)};
\addplot+[ybar, fill=green!60, draw=green!80] coordinates {
(2015,0)(2018,0)(2019,0)(2022,0)(2023,1)(2024,0)(2025,1)};
\legend{Transfer/Distillation, Ensemble, Federated}
\end{axis}
\end{tikzpicture}
\caption{Publication trend of the 15 included empirical studies.}
\label{fig:trend}
\end{figure}
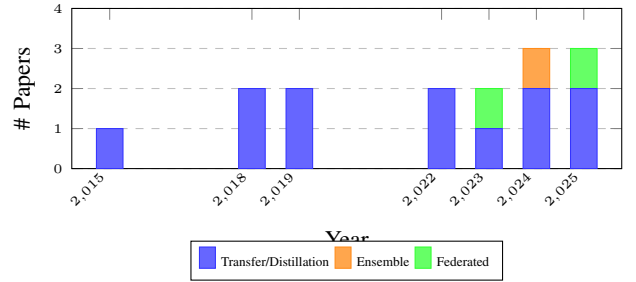

Figure \ref{fig:trend} shows publication timing by mechanism. The corpus is small and recent. Most papers appeared after 2022, but the empirical base is still too thin to support broad mechanism-level rankings.

Table \ref{tab:characteristics} summarizes the included studies.

\begin{table*}[htbp]
\caption{Included empirical studies in the main synthesis (n\,=\,15).}
\label{tab:characteristics}
\renewcommand{\arraystretch}{1.1}
\begin{center}
\small
\begin{tabular}{p{2.1cm}cp{3.0cm}p{3.4cm}p{5.4cm}}
\toprule
\textbf{Reference} & \textbf{Family} & \textbf{Setting} & \textbf{Reuse mechanism} & \textbf{Main reported effect} \\
\midrule
Rusu et al.\ \cite{Rusu2016} & D & Atari & teacher-student KL distillation & Smaller student reaches teacher-level behavior in selected games. \\
Liu et al.\ \cite{Liu2018} & T & cyber-physical control & digital-twin transfer & Synthetic twin data improves learning where real data are scarce. \\
Garc\'{i}a and Fern\'{a}ndez \cite{Garcia2018} & T & safe RL & probabilistic policy reuse & Reuse improves early learning when unsafe exploration matters. \\
Li et al.\ \cite{Li2019} & T & gridworld and PLE & context-aware policy reuse & Selecting when and which policy to reuse improves target learning. \\
Wadhwania et al.\ \cite{Wadhwania2019} & D & multi-agent RL & policy distillation and value matching & Symmetry-aware transfer improves coordination across homogeneous agents. \\
Qu et al.\ \cite{Qu2022} & D & Atari / control & importance-prioritized distillation & Focusing distillation on important frames improves student quality. \\
Zhuang et al.\ \cite{Zhuang2022} & T & skill transfer & skill adaptation and composition & Transfer works when reusable skills are aligned to the target task. \\
Sethi and Pal \cite{Sethi2023} & F & vehicular fog computing & federated parameter averaging & Distributed training reduces raw-data sharing and improves offloading decisions. \\
Zhang et al.\ \cite{Zhang2023} & T & continuous control & Gaussian-process policy reuse & Uncertainty-aware reuse improves sample efficiency over vanilla SAC. \\
Liu et al.\ \cite{Liu2024} & T & policy reuse & Bayesian task inference & Faster task identification reduces early exploration cost. \\
Park et al.\ \cite{Park2024} & E & model-based control & ensemble terminal critics & Probabilistic critic ensembles improve long-horizon value estimates. \\
Yu et al.\ \cite{Yu2024} & D & online distillation & decision-attention distillation & Online teacher guidance helps when informative states are selected adaptively. \\
An et al.\ \cite{An2025} & F & robot navigation & federated RL over ROS/Gazebo & Shared updates improve navigation across distributed robots. \\
Du et al.\ \cite{Du2025} & T & distributed multi-agent control & safe adaptive transfer & Explicit weighting reduces unsafe negative transfer. \\
Sun et al.\ \cite{Sun2025} & D & sparse-reward MARL & leader-collaborator distillation & Distillation supplies denser learning signals for collaborators. \\
\bottomrule
\end{tabular}
\end{center}
\end{table*}

\subsection{Source-target similarity is the clearest recurring condition}
\label{sec:similarity}

The strongest pattern in the corpus is simple: reuse works best when the source and target settings share meaningful structure. That structure takes different forms across papers. In some studies, source and target share underlying dynamics or instrumentation, as in the digital-twin setting of Liu et al.\ \cite{Liu2018}. In others, the shared structure lies in task family or policy semantics, as in context-aware policy reuse \cite{Li2019}, Bayesian policy reuse \cite{Liu2024}, or symmetry-aware multi-agent distillation \cite{Wadhwania2019}.

The converse pattern also appears. When mismatch is large, the reused source can hurt unless the method includes an explicit alignment or gating step. Du et al.\ \cite{Du2025} introduce adaptive weighting to control unsafe transfer in distributed multi-agent settings. Zhang et al.\ \cite{Zhang2023} constrain reuse with Gaussian-process uncertainty, and Li et al.\ \cite{Li2019} learn when to stop reusing a source policy. These papers do not support a universal claim that transfer helps. They support a narrower claim: transfer is more plausible when structural overlap exists and when mismatch is explicitly managed.

\subsection{Evidence for multi-model aggregation is promising but thin}

The literature becomes much thinner once the question shifts from source-to-target transfer to simultaneous use of multiple learned models. Park et al.\ \cite{Park2024} show that an ensemble of terminal critics can improve long-horizon value estimation in model-based control, but that is still one paper in one design family. Federated RL papers \cite{Sethi2023,An2025} show that decentralized aggregation can be useful when privacy or distributed deployment matters, yet both studies focus on relatively homogeneous client settings and do not test strong heterogeneity stress cases.

This matters for interpretation. The current evidence does not justify broad statements such as ``ensembles are the most reliable mechanism'' or ``federated RL is generally beneficial.'' The more defensible reading is narrower: the empirical literature provides several positive demonstrations, but direct cross-family comparison is weak because the evidence base is sparse, tasks differ, and the reported baselines are not standardized.

\subsection{Compute reporting is weak}

The compute argument requires careful qualification. Most included papers do not report training budgets in a way that allows a clean comparison between reused-model methods and stronger from-scratch baselines. Environment steps, wall-clock time, hardware, and pretraining cost are rarely normalized together. In transfer-style papers, the reuse method often benefits from source-task training that is not charged against the target-task baseline. In ensemble and federated papers, multiple networks or clients are trained, but the counterfactual of one stronger model trained with comparable budget is usually not tested.

This does not invalidate the reported gains. It does mean that claims about sample efficiency or compute efficiency should be read as conditional on the reporting choices of the original studies.

\subsection{A provisional independence spectrum}

The included studies suggest a useful way to describe diversity among reused models:
\begin{enumerate}
\item \textbf{Seed diversity}: models differ mainly by random initialization.
\item \textbf{Data diversity}: models see different samples or local client data.
\item \textbf{Task diversity}: models come from different source tasks or roles.
\item \textbf{Reward diversity}: models are trained under different objectives.
\end{enumerate}

The current corpus provides examples of the first three levels. Park et al.\ \cite{Park2024} and the federated papers \cite{Sethi2023,An2025} align most closely with data diversity. Policy reuse and transfer studies often operate at task diversity \cite{Li2019,Liu2024,Du2025}. Reward diversity is largely absent from the final main synthesis. The spectrum is therefore useful as a descriptive language, but it is not yet a validated metric or a proven predictor of outcomes.

\section{Discussion}
\label{sec:discussion}

This synthesis supports three conclusions. First, pretrained knowledge reuse in DRL is not a single mechanism but a family of related interventions whose success depends heavily on compatibility between source and target settings. Second, the available evidence is dominated by transfer and distillation papers, so claims about ensembles, federated training, or direct multi-model aggregation should remain cautious. Third, the literature still reports efficiency claims more aggressively than the underlying compute evidence warrants.

The review also clarifies what should not be claimed. The current evidence does not show that combining models ``consistently'' improves performance across DRL. It does not show that one mechanism family is best in general. It does not support universal statements about compute unfairness, because the studies seldom report enough budget information to audit them rigorously.

\subsection{Limitations}

This review has four important limitations. First, screening and coding were performed by a single reviewer. Second, the search was limited to IEEE Xplore, ACM Digital Library, and citation tracing, which is defensible for engineering-oriented DRL but still incomplete. Third, the exact day-level search log from the original search was not preserved, although the final query structure is reported. Fourth, the final corpus is small and imbalanced: 12 of 15 papers come from transfer or distillation settings, which limits any attempt to rank mechanism families.

\subsection{Implications for future work}

Future empirical work should improve along three axes. It should test reuse methods under clearer source-target mismatch conditions, report stronger baseline and compute controls, and evaluate genuinely multi-model settings beyond narrow case studies. A worthwhile next step would be a benchmark suite that holds target tasks fixed while varying the amount of source-target overlap and the cost assigned to source pretraining. That design would make it possible to test whether the proposed independence spectrum predicts anything beyond intuition.

\section{Conclusion}
\label{sec:conclusions}

This review re-examined the literature on pretrained knowledge reuse in DRL and tightened the evidence base to 15 empirical studies that actually match the review question. The main result is not that reuse always works. The stronger and more defensible result is that positive outcomes are concentrated in settings with substantial structural overlap between source and target tasks or with explicit mechanisms that gate, weight, or align transferred knowledge. Evidence for ensembles and federated aggregation remains promising but too sparse for broad mechanism-level claims, and compute reporting remains too weak for confident statements about efficiency.

The paper's main contribution is therefore a cleaner map of what the current literature does and does not support. The field does not need more inflated claims. It needs compute-controlled baselines, clearer reporting of uncertainty, and direct benchmarks that compare reuse strategies under common task conditions.

\appendices

\section{Query templates}
\label{app:queries}

The review used the following query logic, adapted to database syntax and restricted to title/abstract fields.

\textbf{Intervention block:}
\begin{quote}
``policy distillation'' OR ``knowledge distillation'' OR ``policy reuse'' OR ``policy transfer'' OR ``transfer reinforcement learning'' OR ``ensemble reinforcement learning'' OR ``federated reinforcement learning'' OR ``model merging'' OR ``weight averaging''
\end{quote}

\textbf{RL-context block:}
\begin{quote}
``reinforcement learning'' OR ``deep reinforcement learning'' OR ``actor-critic'' OR ``policy gradient'' OR ``deep Q-network'' OR ``DQN''
\end{quote}

The final database query was the conjunction of the two blocks. Citation tracing was then applied to all papers that survived full-text screening.

\section{Adjacent studies identified but excluded from the main synthesis}
\label{app:adjacent}

\begin{table}[htbp]
\caption{Adjacent studies excluded from the main empirical synthesis.}
\label{tab:excluded_adjacent}
\renewcommand{\arraystretch}{1.12}
\begin{center}
\small
\begin{tabular}{p{2.1cm}p{4.9cm}}
\toprule
\textbf{Reference} & \textbf{Reason for exclusion from main synthesis} \\
\midrule
Zhao et al.\ \cite{Zhao2020} & Survey paper, not a primary empirical DRL study. \\
Wei et al.\ \cite{Wei2025} & Survey paper, not a primary empirical DRL study. \\
Shaik et al.\ \cite{Shaik2025} & Inference-time adversarial filtering rather than reuse of pretrained knowledge during learning. \\
Li et al.\ \cite{Li2025} & Hardware-throughput paper; speeds up RL simulation but does not study model reuse. \\
Piccoli et al.\ \cite{Piccoli2025} & Recent workshop/preprint on feature-level model combination; relevant background but not retained in the archival main synthesis. \\
\bottomrule
\end{tabular}
\end{center}
\end{table}

\section*{Acknowledgment}

The author thanks Prof. Dr. Javad Ghofrani for his guidance and supervision during this research.

\section*{Declaration regarding the Use of Artificial Intelligence}
I used Generative Artificial Intelligence tools during the preparation of this report. 
These tools supported grammar correction, stylistic improvements, and the construction of tables and figures. 
I authored the original technical content and intellectual arguments. 
I reviewed and verified all automated outputs for accuracy. 
I maintain full accountability for the final content.

\balance
\bibliographystyle{IEEEtran}
\bibliography{references}

@article{Mnih2015,
  author    = {Mnih, Volodymyr and Kavukcuoglu, Koray and Silver, David and Rusu, Andrei A. and Veness, Joel and Bellemare, Marc G. and Graves, Alex and Riedmiller, Martin and Fidjeland, Andreas K. and Ostrovski, Georg and Petersen, Stig and Beattie, Charles and Sadik, Amir and Antonoglou, Ioannis and King, Helen and Kumaran, Dharshan and Wierstra, Daan and Legg, Shane and Hassabis, Demis},
  title     = {Human-level control through deep reinforcement learning},
  journal   = {Nature},
  year      = {2015},
  volume    = {518},
  number    = {7540},
  pages     = {529--533},
  doi       = {10.1038/nature14236}
}

@article{Silver2017,
  author    = {Silver, David and Schrittwieser, Julian and Simonyan, Karen and Antonoglou, Ioannis and Huang, Aja and Guez, Arthur and Hubert, Thomas and Baker, Lucas and Lai, Matthew and Bolton, Adrian and Chen, Yutian and Lillicrap, Timothy and Hui, Fan and Sifre, Laurent and van den Driessche, George and Graepel, Thore and Hassabis, Demis},
  title     = {Mastering the game of {Go} without human knowledge},
  journal   = {Nature},
  year      = {2017},
  volume    = {550},
  number    = {7676},
  pages     = {354--359},
  doi       = {10.1038/nature24270}
}

@article{Taylor2009,
  author    = {Taylor, Matthew E. and Stone, Peter},
  title     = {Transfer Learning for Reinforcement Learning Domains: A Survey},
  journal   = {Journal of Machine Learning Research},
  year      = {2009},
  volume    = {10},
  number    = {56},
  pages     = {1633--1685},
  url       = {https://www.jmlr.org/papers/v10/taylor09a.html}
}

@article{Page2021PRISMA,
  author    = {Page, Matthew J. and McKenzie, Joanne E. and Bossuyt, Patrick M. and Boutron, Isabelle and Hoffmann, Tammy C. and Mulrow, Cynthia D. and Shamseer, Larissa and Tetzlaff, Jennifer M. and Akl, Elie A. and Brennan, Sue E. and Chou, Roger and Glanville, Julie and Grimshaw, Jeremy M. and Hróbjartsson, Asbjørn and Lalu, Manoj M. and Li, Tianjing and Loder, Elizabeth W. and Mayo-Wilson, Evan and McDonald, Steve and McGuinness, Luke A. and Stewart, Lesley A. and Thomas, James and Tricco, Andrea C. and Welch, Vivian A. and Whiting, Penny and Moher, David},
  title     = {The {PRISMA} 2020 statement: an updated guideline for reporting systematic reviews},
  journal   = {BMJ},
  year      = {2021},
  volume    = {372},
  pages     = {n71},
  doi       = {10.1136/bmj.n71}
}

@article{Liu2018,
  author    = {Liu, Teng and Tian, Bin and Ai, Yunfeng and Li, Li and Cao, Dongpu and Wang, Fei-Yue},
  title     = {Parallel reinforcement learning: a framework and case study},
  journal   = {IEEE/CAA Journal of Automatica Sinica},
  year      = {2018},
  volume    = {5},
  number    = {4},
  pages     = {827--835},
  doi       = {10.1109/JAS.2018.7511144}
}

@inproceedings{Shaik2025,
  author    = {Shaik, Mohammed Ali and Harshavardhan, B. and Ajay, R. and Rajeev, K.},
  title     = {A Hybrid Ensemble Framework for Adversarial Robustness in Deep Reinforcement Learning},
  booktitle = {2025 6th International Conference on Data Intelligence and Cognitive Informatics (ICDICI)},
  year      = {2025},
  pages     = {1036--1041},
  doi       = {10.1109/ICDICI66477.2025.11135316}
}

@article{Sun2025,
  author    = {Sun, Shaoqi and Liu, Hui and Xu, Kele and Ding, Bo},
  title     = {Leaders and Collaborators: Addressing Sparse Reward Challenges in Multi-Agent Reinforcement Learning},
  journal   = {IEEE Transactions on Emerging Topics in Computational Intelligence},
  year      = {2025},
  volume    = {9},
  number    = {2},
  pages     = {1976--1989},
  doi       = {10.1109/TETCI.2024.3488772}
}

@inproceedings{Li2025,
  author    = {Li, Jiayi and Zhao, Hongxiao and Yue, Wenshuo and Fu, Yihan and Shi, Daijing and Fan, Anjunyi and Yang, Yuchao and Yan, Bonan},
  title     = {{PEARL}: {FPGA}-Based Reinforcement Learning Acceleration with Pipelined Parallel Environments},
  booktitle = {2025 Design, Automation \& Test in Europe Conference (DATE)},
  year      = {2025},
  pages     = {1--7},
  doi       = {10.23919/DATE64628.2025.10992886}
}

@article{Sethi2023,
  author    = {Sethi, Vivek and Pal, Sujata},
  title     = {{FedDOVe}: A Federated Deep {Q}-learning-based Offloading for Vehicular Fog Computing},
  journal   = {Future Generation Computer Systems},
  year      = {2023},
  volume    = {141},
  pages     = {96--105},
  doi       = {10.1016/j.future.2022.11.012}
}

@article{Park2024,
  author    = {Park, Jonghyeok and Jeon, Soo and Han, Soohee},
  title     = {Model-Based Reinforcement Learning With Probabilistic Ensemble Terminal Critics for Data-Efficient Control Applications},
  journal   = {IEEE Transactions on Industrial Electronics},
  year      = {2024},
  volume    = {71},
  number    = {8},
  pages     = {9470--9479},
  doi       = {10.1109/TIE.2023.3331074}
}

@article{An2025,
  author    = {An, Xing and Lin, Yangfei and Lin, Min and Wu, Celimuge and Murase, Tutomu and Ji, Yusheng},
  title     = {Federated Reinforcement Learning Framework for Mobile Robot Navigation Using {ROS} and Gazebo},
  journal   = {IEEE Internet of Things Magazine},
  year      = {2025},
  volume    = {8},
  number    = {5},
  pages     = {45--51},
  doi       = {10.1109/MIOT.2025.3575929}
}

@inproceedings{Zhao2020,
  author    = {Zhao, Wenshuai and Queralta, Jorge Pe\~na and Westerlund, Tomi},
  title     = {Sim-to-Real Transfer in Deep Reinforcement Learning for Robotics: A Survey},
  booktitle = {2020 IEEE Symposium Series on Computational Intelligence (SSCI)},
  year      = {2020},
  pages     = {737--744},
  doi       = {10.1109/SSCI47803.2020.9308468}
}

@inproceedings{Rusu2016,
  author    = {Rusu, Andrei A. and Colmenarejo, Sergio G{\'o}mez and G{\"u}l{\c{c}}ehre, {\c{C}}a{\u{g}}lar and Desjardins, Guillaume and Kirkpatrick, James and Pascanu, Razvan and Mnih, Volodymyr and Kavukcuoglu, Koray and Hadsell, Raia},
  title     = {Policy Distillation},
  booktitle = {4th International Conference on Learning Representations (ICLR), Workshop Track},
  year      = {2016},
  note      = {Initial preprint released in 2015},
  url       = {https://arxiv.org/abs/1511.06295}
}

@misc{Piccoli2025,
  author       = {Piccoli, Elia and Li, Malio and Carf{\`i}, Giacomo and Lomonaco, Vincenzo and Bacciu, Davide},
  title        = {Combining Pre-Trained Models for Enhanced Feature Representation in Reinforcement Learning},
  year         = {2025},
  howpublished = {IBRL @ RLC 2025 workshop paper / preprint},
  url          = {https://openreview.net/forum?id=q8NKvSaLKm}
}

@inproceedings{Wadhwania2019,
  author    = {Wadhwania, Samir and Kim, Dong-Ki and Omidshafiei, Shayegan and How, Jonathan P.},
  title     = {Policy Distillation and Value Matching in Multiagent Reinforcement Learning},
  booktitle = {2019 IEEE/RSJ International Conference on Intelligent Robots and Systems (IROS)},
  year      = {2019},
  pages     = {8193--8200},
  doi       = {10.1109/IROS40897.2019.8967849}
}

@inproceedings{Qu2022,
  author    = {Qu, Xinghua and Ong, Yew Soon and Gupta, Abhishek and Wei, Pengfei and Sun, Zhu and Ma, Zejun},
  title     = {Importance Prioritized Policy Distillation},
  booktitle = {Proceedings of the 28th ACM SIGKDD Conference on Knowledge Discovery and Data Mining},
  year      = {2022},
  pages     = {1420--1429},
  doi       = {10.1145/3534678.3539266}
}

@inproceedings{Li2019,
  author    = {Li, Siyuan and Gu, Fangda and Zhu, Guangxiang and Zhang, Chongjie},
  title     = {Context-Aware Policy Reuse},
  booktitle = {Proceedings of the 18th International Conference on Autonomous Agents and MultiAgent Systems},
  year      = {2019},
  pages     = {989--997},
  url       = {https://dl.acm.org/doi/10.5555/3306127.3331795}
}

@article{Garcia2018,
  author    = {Garc{\'i}a, Javier and Fern{\'a}ndez, Fernando},
  title     = {Probabilistic Policy Reuse for Safe Reinforcement Learning},
  journal   = {ACM Transactions on Autonomous and Adaptive Systems},
  year      = {2018},
  volume    = {13},
  number    = {3},
  pages     = {1--24},
  doi       = {10.1145/3310090}
}

@inproceedings{Zhuang2022,
  author    = {Zhuang, Benhui and Zhang, Chunhong and Hu, Zheng},
  title     = {Policy Transfer via Skill Adaptation and Composition},
  booktitle = {Proceedings of the 2022 6th International Conference on Computer Science and Artificial Intelligence},
  year      = {2022},
  pages     = {195--202},
  doi       = {10.1145/3577530.3577562}
}

@inproceedings{Yu2024,
  author    = {Yu, Xinqiang and Yang, Chuanguang and Yu, Chengqing and Huang, Libo and An, Zhulin and Xu, Yongjun},
  title     = {Online Policy Distillation with Decision-Attention},
  booktitle = {2024 International Joint Conference on Neural Networks (IJCNN)},
  year      = {2024},
  pages     = {1--8},
  doi       = {10.1109/IJCNN60899.2024.10651412}
}

@article{Du2025,
  author    = {Du, Bin and Xie, Wei and Li, Yang and Yang, Qisong and Zhang, Weidong and Negenborn, Rudy R. and Pang, Yusong and Chen, Hongtian},
  title     = {Safe Adaptive Policy Transfer Reinforcement Learning for Distributed Multiagent Control},
  journal   = {IEEE Transactions on Neural Networks and Learning Systems},
  year      = {2025},
  volume    = {36},
  number    = {1},
  pages     = {1939--1946},
  doi       = {10.1109/TNNLS.2023.3326867}
}

@article{Liu2024,
  author    = {Liu, Jinmei and Wang, Zhi and Chen, Chunlin and Dong, Daoyi},
  title     = {Efficient Bayesian Policy Reuse With a Scalable Observation Model in Deep Reinforcement Learning},
  journal   = {IEEE Transactions on Neural Networks and Learning Systems},
  year      = {2024},
  volume    = {35},
  number    = {10},
  pages     = {14797--14809},
  doi       = {10.1109/TNNLS.2023.3281604}
}

@inproceedings{Zhang2023,
  author    = {Zhang, Wenzhuo and Tang, Tao and Cui, Jiabao and Liu, Shuang and Xu, Xin},
  title     = {Transfer Reinforcement Learning Based on Gaussian Process Policy Reuse},
  booktitle = {2023 7th Asian Conference on Artificial Intelligence Technology (ACAIT)},
  year      = {2023},
  pages     = {1491--1500},
  doi       = {10.1109/ACAIT60137.2023.10528459}
}

@inproceedings{Wei2025,
  author    = {Wei, Jiaye and Lan, Yixing and Tang, Tao and Liu, Tenglong},
  title     = {A Survey on Transfer Reinforcement Learning},
  booktitle = {2025 8th International Conference on Advanced Algorithms and Control Engineering (ICAACE)},
  year      = {2025},
  pages     = {2511--2518},
  doi       = {10.1109/ICAACE65325.2025.11020301}
}

\end{document}